\documentclass[11pt]{article}

\usepackage[a4paper,margin=1in]{geometry}
\usepackage[T1]{fontenc}
\usepackage[utf8]{inputenc}
\usepackage{graphicx}
\usepackage{booktabs}
\usepackage{multirow}
\usepackage{array}
\usepackage{amsmath}
\usepackage{amssymb}
\usepackage{caption}
\usepackage{subcaption}
\usepackage{url}
\usepackage{hyperref}
\hypersetup{colorlinks=true, linkcolor=blue, citecolor=blue, urlcolor=blue}

\providecommand{\Description}[1]{}

\begin{document}

\begin{center}

{\LARGE\bfseries NEST3D: A High-Resolution Multimodal Dataset and\\[0.2em]
Benchmark for Sociable Weaver Tree Nests}

\vspace{0.9em}

{\normalsize
Constanza A. Molina Catricheo$^{1}$,
Simon Boeder$^{1}$,
Ting-Jia Guo$^{1}$,
Giacomo May$^{2}$,\\[0.25em]
Clément Berthelot$^{3,4}$,
Devis Tuia$^{2}$,
Friedrich F. Reinhard$^{5}$,
Fabio Remondino$^{6}$,
Benjamin Risse$^{1}$
}

\vspace{0.6em}

{\small
$^{1}$Institute for Geoinformatics (ifgi), University of Münster, Germany\\
$^{2}$École Polytechnique Fédérale de Lausanne (EPFL), Switzerland\\
$^{3}$Max Planck Institute of Animal Behavior, Germany\\
$^{4}$University of Konstanz, Germany\\
$^{5}$Kuzikus Research Station, Namibia\\
$^{6}$Fondazione Bruno Kessler (FBK), Italy\\[0.35em]
\texttt{cmolinac@uni-muenster.de}
}

\end{center}

\vspace{0.5em}

\begin{abstract}
Sociable weaver nests function as complex ecological structures offering thermoregulatory microhabitats and sustaining diverse species; however, datasets used in prior studies lack fine-grained 3D structural detail.
Producing usable and accurate 3D weaver nest data is challenging due to their irregular geometry and integration with complex host vegetation.
We bridge this gap with an open-access, 1.4 TB multimodal drone dataset of 104 nest-bearing trees, comprising 27,945 RGB images, 111,780 multispectral images, approximately 781 million 3D points, and expert-annotated semantic segmentation labels.
We benchmark semantic segmentation using KPConv, RandLA-Net, and Point Transformer V3, with PT-v3 achieving an mIoU of 86.35\% on the test set.
While the results demonstrate strong performance for transformer-based and point-wise methods, they also highlight architecture-dependent challenges, particularly for convolution-based approaches such as KPConv.
By uniquely combining spectral, spatial, and structural information, the presented dataset advances 3D reconstruction, segmentation, and classification algorithms, enabling ecological applications from nest volume estimation to species conservation, and serves as a demanding benchmark that exposes architecture-dependent performance under extreme class imbalance.
\end{abstract}


\begin{figure}[h!]
\centering
\includegraphics[width=\textwidth]{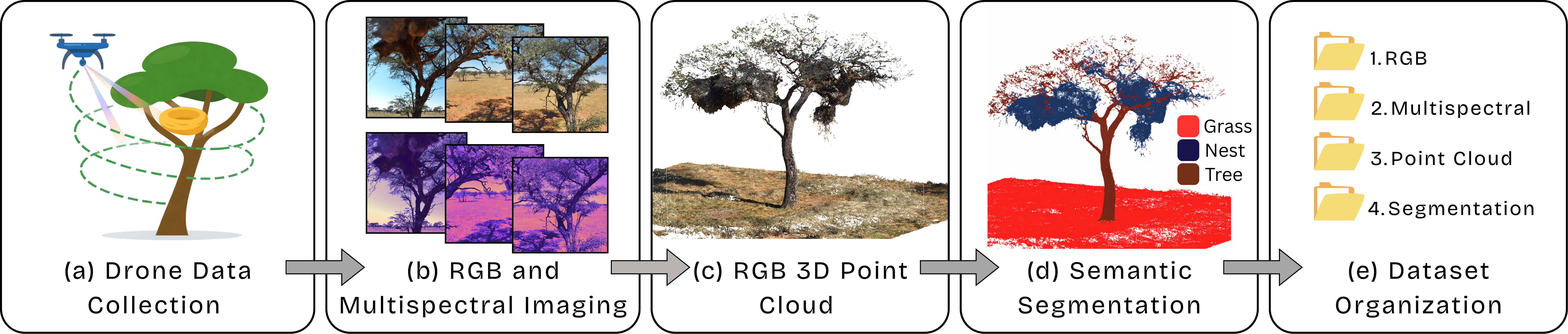}
\caption{Workflow for creating the NEST3D dataset:
(a) UAV-based data collection around a single tree,
(b) RGB and corresponding multispectral images,
(c) RGB 3D point cloud reconstruction,
(d) semantic segmentation highlighting grass (red), nest (dark blue), and tree (brown),
(e) organized multimodal dataset folders.}
\label{fig:workflow}
\end{figure}

\vspace{1em}
\section{Introduction}

Sociable weavers (\textit{Philetairus socius}) build large, unique communal nests that serve as some of the largest known structures created by birds, housing over 500 individuals and persisting for several decades across many generations \cite{maclean1973sociable, COLLIAS1978}.
These impressive structures function as complex ecological systems, offering thermoregulatory benefits by buffering temperature extremes and creating localized biodiversity hotspots for invertebrates, reptiles, birds, and mammals \cite{perez2020climate, lowney2022ecological, 10.3389fevo.2020.570006}.
While prior measurements of individual nest volumes have ranged from $0.7$ to $10\text{ m}^3$ \cite{van2013thermoregulatory}, high-resolution multimodal datasets capturing the precise three-dimensional structural detail of these colonies remain scarce \cite{bartholomew1976thermal}.

Remote sensing with Unmanned Aerial Vehicles (UAVs) has revolutionized ecological monitoring by enabling non-invasive, high-speed surveys of diverse ecological features \cite{delplanque2024will, linchant2015unmanned}.
Drones allow for the detailed mapping of vegetation cover, the spatial distribution of nesting sites, and the topography of the savanna landscape, which are critical for assessing habitat health and biodiversity \cite{robinson2022existing, jing2021multispectral}.
These capabilities are invaluable for assessing vegetation health, habitat structure, and biodiversity, especially in challenging terrains like the arid savannas where sociable weavers reside.
Despite the recognized ecological significance of sociable weaver nests \cite{bartholomew1976thermal}, high-resolution, multimodal datasets capturing their structural complexity are entirely absent, leaving a critical gap for ecological research.

\begin{figure}[ht]
\centering
\includegraphics[width=1.0\linewidth]{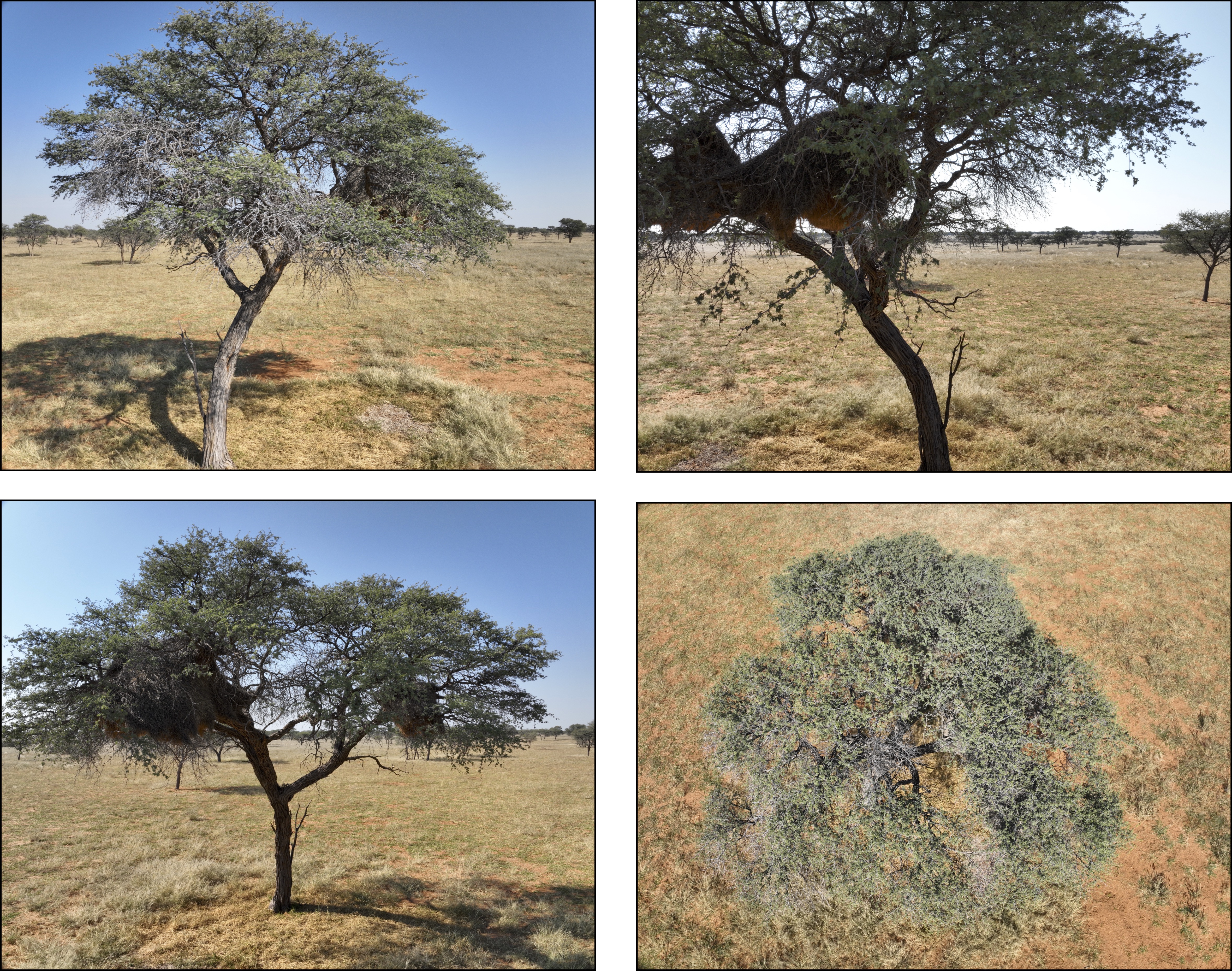}
\caption{Example tree-nest structures from the dataset. These four RGB images illustrate the diverse geometries, varying levels of occlusion, and the complex visual interface between the nests and the dense tree canopy.}
\label{fig:nest_example}
\end{figure}

Automating the analysis of these systems presents significant computational challenges.
From a computer vision perspective, the 3D reconstruction of complex vegetation is notoriously difficult due to severe self-occlusion, thin branch structures, and a lack of planar surfaces \cite{okura20223d}.
Furthermore, semantic segmentation is non-trivial: the nests possess irregular, amorphous geometries that often appear spectrally indistinguishable from the woody biomass of the host tree in the visible spectrum (Figure~\ref{fig:nest_example}).
Robustly disentangling the nest from the supporting canopy requires data-driven approaches capable of learning subtle structural and textural features, yet no such benchmark datasets currently exist.

In this work, we present \textbf{NEST3D}, an open multi-modal dataset of sociable weaver nests (an overview is given in Figure~\ref{fig:workflow}).

Our contributions are summarized as follows:
\begin{itemize}
    \item \textbf{Multimodal Dataset:} We provide a high-resolution resource consisting of 104 nest-bearing trees, including RGB imagery, four-band multispectral data (Green, Red, Red Edge, and NIR), and 3D RGB point clouds.
    \item \textbf{Expert Annotations:} We release dense 3D point clouds manually annotated by experts into three semantic classes (\textit{tree}, \textit{nest}, and \textit{grass}), providing a ground truth for scene understanding.
    \item \textbf{Benchmarking SOTA Models:} We evaluate three state-of-the-art 3D architectures (PT-v3, RandLA-Net, and KPConv), establishing baseline results that highlight the challenges of extreme class imbalance in ecological remote sensing.
    \item \textbf{Cross-disciplinary Application:} This resource bridges the gap between field ecology and computational analysis \cite{tuia2022perspectives}, enabling applications such as automated nest volume estimation and vegetation health analysis.
\end{itemize}

\section{Related Work}

The study of colonial nesting structures includes ecology, remote sensing, and computer vision.
Sociable weaver nests in particular have been widely documented in biological literature, with emphasis on social organization, thermoregulation, colony longevity, and ecological interactions \cite{COLLIAS1978, van2013thermoregulatory, maclean1973sociable}.
These studies are primarily based on field observations, manual measurements, and colony-level summaries.
While they provide critical biological insight, they do not offer spatially explicit digital representations of nest geometry, nor publicly available annotated data suitable for computational analysis or algorithmic benchmarking.

Recent progress in UAV-based remote sensing has enabled detailed mapping of vegetation and individual trees using high-resolution RGB imagery, multispectral sensors, and LiDAR laser.
Existing datasets and methods focus on tasks such as tree crown delineation, vegetation classification, biomass estimation, and canopy segmentation \cite{spiers2025review}.
These efforts demonstrate the value of aerial data for ecological monitoring, but they generally treat nests as part of the surrounding vegetation \cite{ouaknine2025openforest}.
Among the few studies that explicitly attempt nest detection \cite{andrew2017semi}, the nests are built on the ground, on low shrubs, or only occasionally within the sparse canopy, showing that automated pipelines have not yet addressed complex, canopy-embedded nests such as those of sociable weavers.
Moreover, most released datasets either operate in two dimensions or provide unannotated 3D reconstructions, limiting their use for supervised learning on fine-grained structural tasks \cite{ouaknine2025openforest}.

\subsection{Sociable Weaver Nests and Tree Ecology}

Found predominantly in the arid and semi-arid savannas of southern Africa, sociable weavers rely heavily on trees---most commonly \textit{Vachellia erioloba} (camel thorn)---to support their large nests, which can weigh several hundred kilograms.

Studies cover four key areas: (i) cooperative behavior in nest construction \cite{leighton2014sex, leighton2016sociable, COLLIAS1978}; (ii) thermoregulation via chamber organization \cite{White1975, bartholomew1976thermal, 10.3389fevo.2020.570006, SilvaBenoit2024}; (iii) material selection and century-scale durability \cite{SilvaBenoit2024}; (iv) biodiversity hotspot roles \cite{lowney2022ecological}, among others.

Despite their ecological importance, understanding of the structural and spatial characteristics of sociable weaver nests remains limited to field-based measurements such as nest size, chamber counts, or branch distribution.
These measurements cannot capture 3D geometry or nest-tree spatial relationships, which are critical for analyzing structural constraints and microclimates, and no publicly available dataset currently provides annotated 3D nest representations suitable for training or benchmarking data segmentation models in this ecologically important domain.

\subsection{3D Tree Reconstruction Datasets}

Three dimensional reconstructions of natural habitats have become essential for ecosystem analysis, from canopy height models to landscape topography \cite{10.1093/biosci/biaf069}.
These models are typically generated using terrestrial laser scanning (TLS), drone, aerial-based LiDAR, and photogrammetry \cite{scher2019drone, hartley2024tree}.

Notable recent works range from extraction of individual tree measurements in Eucalyptus plantations using drone-LiDAR \cite{zhou2025extraction} to simulation ready tree datasets generated from single images with diffusion priors \cite{lee2024tree}.
In addition, multispectral LiDAR 3D data has been used to benchmark 3D point cloud semantic segmentation, demonstrating that multispectral information improves segmentation performance \cite{takhtkeshha20253d}.

While these datasets provide valuable resources for tree detection and canopy analysis, they generally treat animal-built structures such as sociable weaver nests as part of the surrounding vegetation.
None include nest-specific semantic labels or capture nest-tree spatial relationships, both of which are critical for fine-grained ecological studies.
To our knowledge, no public 3D dataset offers annotated sociable weaver nests as distinct classes separate from vegetation, highlighting a clear gap that NEST3D addresses.

\section{Methodology}

\subsection{Dataset Organization and Content}

We present a multimodal dataset of sociable weaver nests, collected using drone-based RGB and multispectral imagery and reconstructed into semantically annotated 3D point clouds for benchmarking scene-level semantic segmentation.
The dataset combines RGB imagery, multispectral data, and 3D spatial representations of trees hosting sociable weaver nests, with point-level semantic annotations.

The dataset is publicly available through the Hugging Face Hub\footnote{\url{https://doi.org/10.57967/hf/8978}} and is organized into modality-specific directories to support flexible access and reuse.
The data are split into fixed \texttt{train} and \texttt{test} sets, each containing a consistent collection of scenes.

\begin{itemize}
\item \textbf{RGB imagery:} Raw drone images organized by data split (\texttt{train}, \texttt{test}) and tree--nest scene (e.g., \path{train/sample_012/RGB/sample012_RGB_119.JPG}).
\item \textbf{Multispectral imagery (four directories inside \texttt{MS}):} Images from the same acquisitions as RGB, organized by split and scene into four band-specific folders (Green, Red, Red Edge, NIR) within \texttt{MS} (e.g., \path{train/sample_012/MS/Green/sample012_G_119.TIF}).
\item \textbf{3D point clouds (.npy files):} One NumPy file per scene (e.g., \path{train/sample012/sample012.npy}) containing per-point attributes \texttt{[x, y, z, r, g, b, label]}.
\end{itemize}

\subsection{Acquisition Site}

The study is conducted at the Kuzikus Wildlife Reserve, located in the Kalahari Desert of Namibia (approximately 23\textdegree{}S, 18\textdegree{}E).
The region exhibits a semi-arid climate characterized by hot summers, with daytime temperatures frequently exceeding 40\,\textdegree{}C, and cold, dry winters during which nighttime temperatures can approach 0\,\textdegree{}C.

Annual rainfall averages 220--240\,mm and occurs predominantly between December and March \cite{shikangalah2022responsiveness}.
The study area encompasses the entire Kuzikus Wildlife Reserve, covering approximately 100\,km$^{2}$.
This protected area supports a diverse assemblage of Kalahari fauna, including more than 40 mammal species and over 100 bird species as stated in the Kuzikus internal species inventory.

Sociable weaver colonies occur exclusively in the dominant tree species of the region, particularly the camel-thorn (\textit{Vachellia erioloba}) and the shepherd's tree (\textit{Boscia albitrunca}).
These species provide structurally suitable nesting sites within the otherwise open savanna landscape, shaping the spatial distribution of colonies across the reserve \cite{kuzikus2023}.

\subsection{Data Acquisition}

The dataset is acquired using a DJI Mavic~3~Multispectral drone, equipped with a high-resolution 20\,MP RGB camera and a multispectral camera comprising four 5\,MP sensors that capture green (560\,$\pm$\,16\,nm), red (650\,$\pm$\,16\,nm), red edge (730\,$\pm$\,16\,nm), and near-infrared (860\,$\pm$\,26\,nm) spectral bands.

The multispectral imaging system features a 1/2.8-inch CMOS sensor with electronic shutter speeds ranging from 1/30 to 1/12800 seconds, enabling high-quality image acquisition with minimal motion blur.
The RGB camera is equipped with a 4/3 CMOS sensor and a mechanical shutter capable of speeds up to 1/2000 seconds, supporting fast burst modes and detailed visual capture.
Field data collection takes place from May~30 to June~22,~2025, with two daily flight sessions conducted in the morning (starting at 8:30\,AM) and in the afternoon (starting at 3:30\,PM).

Flights are performed by manually circling individual trees while maintaining an approximate distance of 5--10\,m from the canopy, with occasional increases in distance to avoid disturbing the birds.
This standoff distance balances the need to minimize occlusions with the requirement to maximize nest visibility.
Environmental conditions typical of the semi-arid study area are considered to optimize illumination, although some acquisitions occur near sunset, introducing lower-light conditions.
The integrated sunlight sensor and advanced radiometric capabilities of the DJI Mavic~3~Multispectral help compensate for illumination variability, ensuring consistent image quality throughout the acquisition period.

\subsection{Imagery}

As summarized in \autoref{tab:dataset_stats}, the dataset comprises 104 scenes (83 train, 21 test), totaling 23,139 RGB images for training and 4,806 for testing; the number of RGB images per tree ranges from 111 to 573.
For each sample, RGB and multispectral images were acquired simultaneously during the same flight pass, with multiple passes ensuring comprehensive coverage of all viewing angles.
Each tree corresponds to a single sample in the dataset, and no trees are duplicated across splits.

\subsection{3D Reconstruction}

We generate 3D point clouds using photogrammetry in Agisoft Metashape~\cite{metashape}, relying exclusively on RGB images.
Image orientation is performed iteratively, followed by dense reconstruction from depth maps to produce a dense point cloud for each tree.

The reconstruction is scaled to metric units using the drone GPS coordinates, ensuring that distances and dimensions in the 3D models correspond to real-world measurements.
We note, however, that these measurements inherit the intrinsic positional error of the GPS system, which can affect absolute accuracy at the centimeter level.

During orientation, we set key point limits to 10{,}000 and tie point limits between 20{,}000 and 40{,}000 to achieve a balance between computational efficiency and reconstruction completeness.
We constrain reprojection error to below 2.0\,px and require triangulation angles greater than 2--5\textdegree{} to promote stable geometry.
Points observed in fewer than three images and those within the highest 20--50\% uncertainty range are removed to improve accuracy.
These thresholds are selected to provide an optimal compromise between point cloud density and reconstruction quality based on empirical testing.

Post-processing includes manual inspection and the removal of noise and artifacts to further improve structural accuracy.
This filtering ensures that subsequent analyses are performed on precise and relevant point cloud data.

\subsection{Semantic Segmentation}

The dense point clouds are manually annotated by experts using CloudCompare~\cite{CloudCompare}, where each point is assigned a semantic label corresponding to one of the following three classes: tree, nest, or grass.
Annotators use CloudCompare's selection and segmentation tools to identify and label points, ensuring that the resulting annotations accurately reflect the structures observed in the 3D reconstructions.
\autoref{tab:dataset_stats} summarizes descriptive statistics over all scenes, including the number of RGB and multispectral images, the total point counts, and the relative proportion of points assigned to each semantic class.

\begin{table}[ht]
\centering
\caption{Summary statistics of NEST3D dataset. Point counts in millions ($10^6$). Nest class $<5\%$. Multispectral images = 4$\times$RGB per scene.}
\label{tab:dataset_stats}
\begin{tabular}{lcc}
\toprule
\textbf{Statistic}           & \textbf{Train Set}  & \textbf{Test Set}  \\ \midrule
Number of Scenes             & 83                  & 21                 \\
Total Points ($\times 10^6$) & 686.57              & 94.94              \\
Mean Points per Scene        & $8.27 \pm 4.62$M    & $4.52 \pm 1.92$M   \\
Min / Max Points             & 1.05M / 25.43M      & 1.89M / 8.78M      \\
RGB Images                   & 23,139              & 4,806              \\
\midrule
\textbf{Class Distribution}  & \textbf{Percentage} & \textbf{Percentage} \\ \midrule
Grass                        & 61.70\%             & 49.19\%            \\
Tree                         & 33.91\%             & 45.99\%            \\
Nest                         & 4.39\%              & 4.82\%             \\ \bottomrule
\end{tabular}
\end{table}

\subsection{Benchmark and Evaluation Protocol}
\label{sec:benchmark_protocol}

We consider 3D semantic segmentation on sample-level point clouds with three classes: grass, nest, and tree.
The dataset is split into an 80/20 train/test partition over the 104 samples, resulting in 83 training scenes and 21 test scenes.
Both splits are provided as separate folders on Hugging Face, with identical scene identifiers and modality structure.

We evaluate the manual annotations using three state-of-the-art 3D point cloud semantic segmentation architectures on the proposed benchmark: Point Transformer V3 (PT-v3)~\cite{wu2024ptv3}, RandLA-Net~\cite{hu2020randlanet}, and KPConv~\cite{thomas2019kpconv}.
These models are selected to represent diverse approaches in 3D segmentation, including convolution-based methods (KPConv), sparse point sampling networks (RandLA-Net), and transformer-based architectures (PT-v3), providing a comprehensive assessment of current methods on the benchmark.
Each model is trained and tested on the corresponding point clouds according to its standard pipeline.

All models train and evaluate on the 3D point clouds in NumPy format, where each point is represented as $[x, y, z, r, g, b, l]$, with semantic label $l \in \{ \text{tree}, \text{nest}, \text{grass} \}$.
The samples are expressed in a local Euclidean frame (meters) and normalized into the unit sphere; RGB values are scaled to $[0, 1]$.
Any additional internal preprocessing is performed according to each model's requirements.

For each method, we train on the training split and select hyperparameters using a subset of the training scenes as validation.
We report standard segmentation metrics, including overall accuracy (OA), mean Intersection over Union (mIoU), and mean class accuracy (mAcc).
Detailed per-class metrics and per-scene results are provided in the supplementary material for transparency and reproducibility.

\section{Experiments and Results}

All experiments follow the benchmark protocol described in \autoref{sec:benchmark_protocol}, using a fixed train/test split, identical input features ($XYZ + RGB$), and unit-sphere normalization.

\subsection{Implementation Details}

We use the official implementation of PT-v3\footnote{\url{https://github.com/Pointcept/PointTransformerV3}} and the Open3D-ML implementations of RandLA-Net and KPConv\footnote{\url{https://github.com/isl-org/Open3D-ML}}.

\paragraph{Point Transformer V3 (PT-v3).}
We employ the PT-v3-m1 backbone, consisting of five encoder and four decoder stages (channel widths 32--512).
Training utilizes the AdamW optimizer with a OneCycleLR schedule.
To mitigate the extreme class imbalance, we combine class-weighted cross-entropy ($4\times$ weight for the nest class) with Lovász loss.
Input clouds are voxelized at a 0.008 grid size and randomly cropped to 50k points.
The model is trained for 100 epochs and the final checkpoint is evaluated on the test split.

\paragraph{RandLA-Net.}
The model consists of three layers aggregating information from 32 nearest neighbors.
Subsampling ratios are [1, 1, 2].
We use a class-balanced random sampler with weights [1, 3, 1] during training.
The model is trained for 100 epochs using the Adam optimizer with a learning rate of 0.001 and a batch size of 4.

\paragraph{KPConv.}
The KPConv architecture is configured with a first subsampling distance of 0.01, an input radius of 0.2, and a convolution radius of 0.08.
Each of the three convolutional layers uses 50 kernel points.
Training is performed for 100 epochs using the Adam optimizer (LR: 0.001, Batch: 6) with an exponential decay scheduler.

\subsection{Quantitative Results}

\begin{figure}[t]
    \centering
    \includegraphics[width=0.95\textwidth]{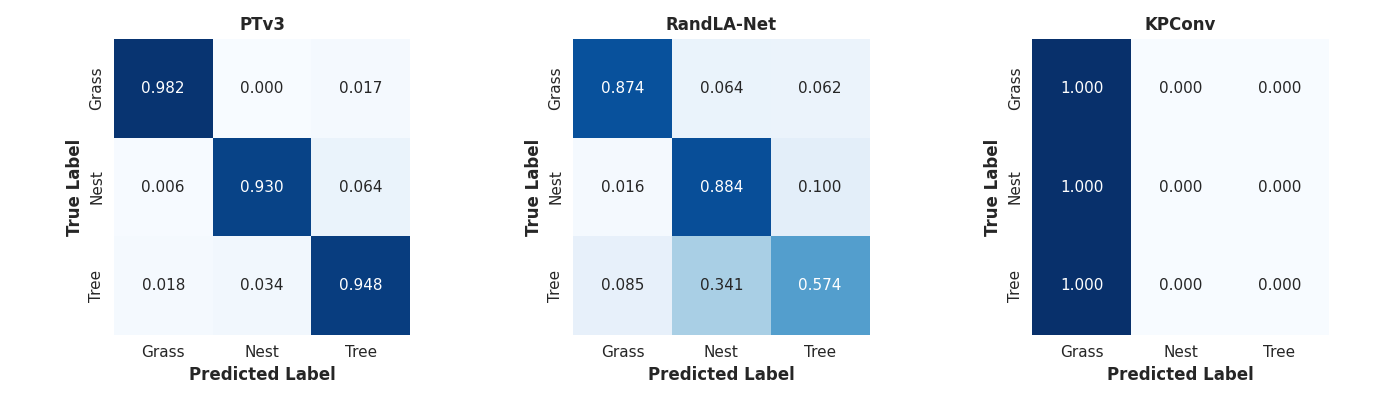}
    \caption{Confusion matrices for 3D point cloud classification comparing PTv3, RandLA-Net, and KPConv models. Each cell shows the normalized proportion of predictions, where darker blue indicates higher values. Note that KPConv failed to learn meaningful structures (for details see text).}
    \label{fig:confusion_matrices}
\end{figure}

\autoref{tab:nest_results} summarizes the performance on the test set, and \autoref{fig:confusion_matrices} presents the corresponding confusion matrices for deeper analysis.
\textbf{PT-v3} achieves the highest overall performance with an OA of 96.42\% and an mIoU of 86.35\%.
Notably, it maintains a \textbf{Nest IoU of 69.99\%} and maintains high true positive rates across all classes (above 0.93).
However, slight confusion is observed between nest and tree points which is expected given the strong visual and structural continuity at the tree--nest interface.
These results demonstrate its ability to handle severe class imbalance by leveraging global and local self-attention mechanisms to better isolate nest geometry from the dense canopy.

In contrast, \textbf{RandLA-Net} performs competitively on majority classes (Grass/Tree) but shows a significant drop in \textbf{Nest IoU (17.98\%)}.
The confusion matrix reveals that while the model identifies 88.4\% of actual Nest points, it suffers from a high false positive rate; specifically, it misclassifies 34.1\% of Tree points as Nests (cf.\ \autoref{fig:confusion_matrices}).
This reflects its sensitivity to limited minority class points and the difficulty of capturing fine-grained structures with random sampling aggregation.
\textbf{KPConv} exhibits a total lack of generalization to minority classes (IoU=0.00\%), suggesting that the model easily converges to a local minimum during training.
As shown in \autoref{fig:confusion_matrices}, the model has completely collapsed to the majority class, predicting ``Grass'' for 100\% of all input points.
Consequently, its OA (49.19\%) is simply a reflection of the dataset's class distribution rather than learned discriminative features.
This lack of recovery indicates that the sparse and non-uniform point distributions inherent in drone-acquired ecological data pose a significant challenge for the stability of kernel-based convolutional gradients.

\begin{table}[ht]
\centering
\caption{Quantitative evaluation on the test set. Per-class Intersection over Union (IoU), mean IoU (mIoU), and Overall Accuracy (OA) are reported as percentages (\%). Best results are highlighted in \textbf{bold}.}
\label{tab:nest_results}
\begin{tabular}{l ccc cc}
\toprule
\textbf{Method} & \textbf{Grass} & \textbf{Nest} & \textbf{Tree} & \textbf{mIoU} & \textbf{OA} \\
                & \textbf{IoU}   & \textbf{IoU}  & \textbf{IoU}  &               &             \\ \midrule
PT-v3           & \textbf{96.59} & \textbf{69.99} & \textbf{92.47} & \textbf{86.35} & \textbf{96.42} \\
RandLA-Net      & 80.86          & 17.98          & 53.30          & 50.72          & 73.64          \\
KPConv          & 49.19          & 0.00           & 0.00           & 16.40          & 49.19          \\ \bottomrule
\end{tabular}
\end{table}

\subsection{Qualitative Analysis}

\begin{figure}[ht]
\centering
\includegraphics[width=1.0\linewidth]{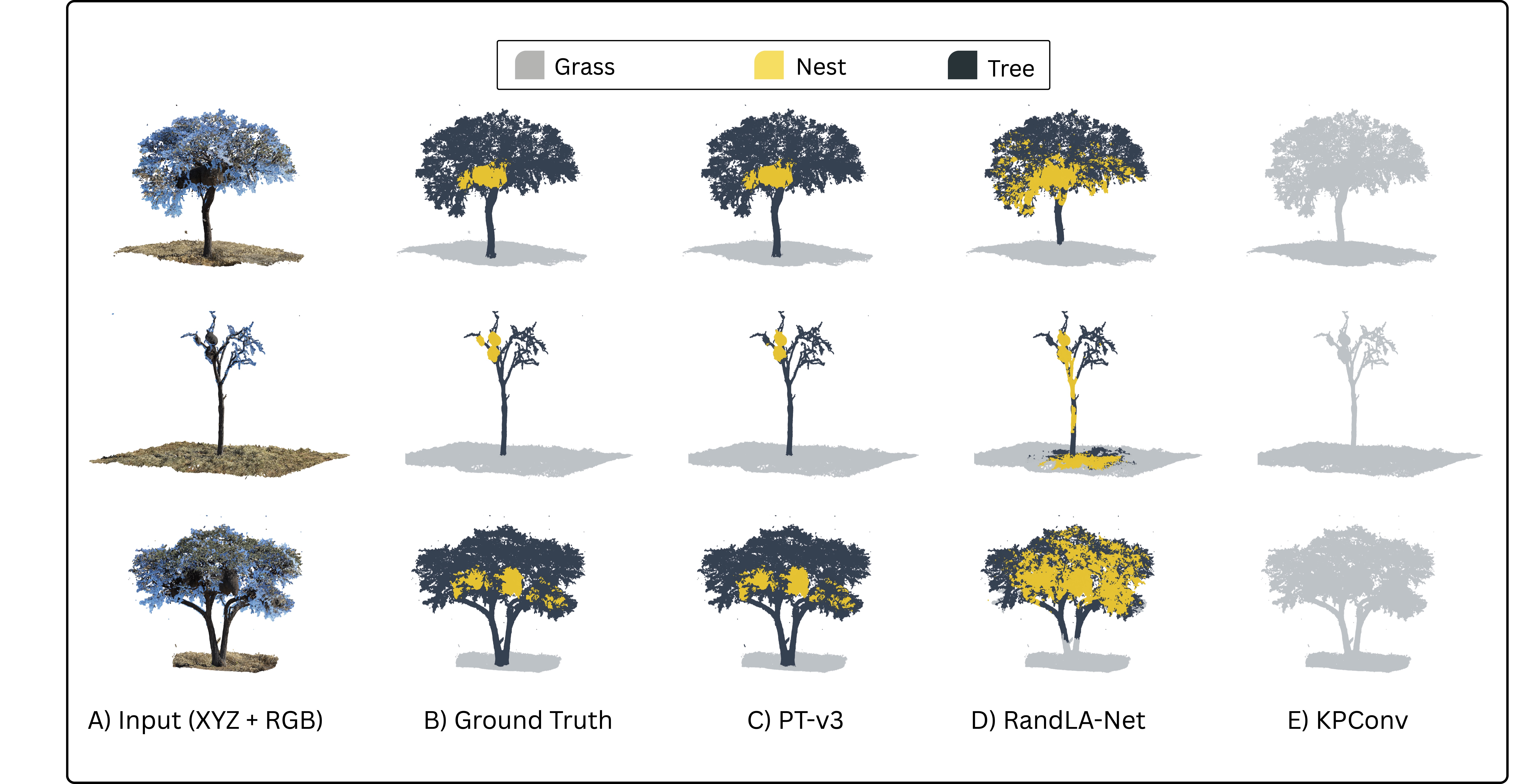}
\caption{Qualitative comparison of 3D semantic segmentation architectures. Column (A) displays the input point clouds ($XYZ+RGB$). Column (B) shows ground truth labels. Columns (C--E) compare model predictions. The color scheme highlights the \textbf{Nest} class (yellow) against the \textbf{Tree} (dark navy) and \textbf{Grass} (light grey) backgrounds. PT-v3 demonstrates superior precision, while RandLA-Net suffers from over-segmentation.}
\label{fig:predictions}
\end{figure}

As illustrated in \autoref{fig:predictions}, the visual results reinforce the numerical findings.
While \textbf{PT-v3} closely aligns with the ground truth, capturing the geometric nuances of the nests, \textbf{RandLA-Net} struggles with boundary definition.
It exhibits significant vertical ``leakage,'' where tree branches and sections of the trunk are misclassified as part of the nest structure.

The failure of \textbf{KPConv} is visually evident as it defaults to the majority ``Grass'' class (light grey) across all samples.
These results highlight a research gap: while transformer-based models handle imbalances better, accurately segmenting small, complex structures in 3D remains an open problem.
Future work may explore specialized sampling or multimodal input (e.g., multispectral data) to improve the segmentation of fine-scale classes.

\section{Ethics, Fairness, and Limitations}

Data collection was conducted with official permission from the owners of the Kuzikus Wildlife Reserve and in strict accordance with local regulations and institutional guidelines for drone-based ecological fieldwork.
As part of the Black Rhino Custodian Project, Kuzikus is actively engaged in safeguarding resident black rhino populations from poaching and other threats, so particular care was taken to avoid capturing imagery of rhinos.
Similarly, no human subjects appear in the dataset except for the authors, who explicitly consent to being visible in the images.

\section{Conclusion}

NEST3D introduces a new multi-modal dataset including RGB, multispectral, and 3D point cloud reconstruction with 3D semantic segmentation annotations of 3 classes: grass, tree and nest.
To assess the quality of the manual segmentation, three baseline semantic segmentation models (PT-v3, RandLA-Net, KPConv) were evaluated under a unified training and testing protocol.
In these experiments, we focus only on RGB and 3D point cloud inputs, leaving the integration of multispectral data as an open avenue for future research.

On the ecological side, one can fuse 3D geometry with multispectral imagery to build multispectral 3D canopy models, enabling canopy segmentation and the computation of vegetation indices to localize stress patterns around and within nest-bearing trees.
These indices could be correlated with nest positions to explore whether sociable weavers preferentially occupy healthier trees or whether nest load and placement affect tree vitality.
At the structural level, the dataset enables quantitative studies that relate nest geometry to branch architecture, or investigate spatial patterns of nest placement within individual trees.
Overall, NEST3D provides the first high-resolution, multimodal resource capturing the structural and spectral complexity of sociable weaver colonies, offering unprecedented opportunities for both ecological analysis and methodological development.

On the computer vision side, the dataset is suitable for cross-dataset generalization experiments, for example training on existing forest or habitat point cloud datasets and testing on this benchmark, or augmenting other point cloud benchmarks with a ``nest'' category to probe transfer to rare ecological classes.
This setting naturally raises questions about how current 3D architectures generalize across ecological domains, and how strong class imbalance and long-tail label distributions affect performance, calibration, and uncertainty estimates.
Overall, this benchmark aims to bring together computer vision and ecology, encouraging interdisciplinary work on complex natural environments and supporting researchers to learn from each other and push this area forward together.
By combining detailed 3D structural data with semantic labels and multispectral imagery, NEST3D establishes a challenging benchmark for segmentation, classification, and 3D reconstruction tasks, encouraging the development of models that can handle highly imbalanced and complex ecological scenes.
Finally, the experimental evaluation of KPConv, RandLA-Net, and PT-v3 serves as a reference baseline, illustrating both the promise and limitations of current architectures on this dataset.

\section*{Acknowledgments}

This work was funded by the European Union's Horizon Europe research and innovation programme through the Marie Sk\l{}odowska-Curie project ``WildDrone -- Autonomous Drones for Nature Conservation'' (grant agreement no.\ 101071224), the EPSRC-funded Autonomous Drones for Nature Conservation Missions grant (EP/X029077/1), and the Swiss State Secretariat for Education, Research and Innovation (SERI) under contract number 22.00280.

We thank our collaborators and field partners in Namibia for their invaluable support during data collection.

\section*{Data Availability}

The NEST3D dataset is publicly available on Hugging Face under a CC BY 4.0 license at \url{https://doi.org/10.57967/hf/8978}.
Version 1.0 was used for all experiments reported in this paper.
The repository contains RGB and multispectral imagery, 3D point clouds, and per-point semantic segmentation labels for all 104 trees.

\bibliographystyle{unsrt}
\bibliography{references}

\clearpage
\appendix
\section*{Supplementary Material}
\label{sec:supplementary}

\subsection*{Detailed Evaluation Metrics}

For each test scene, we report overall accuracy (OA), mean Intersection over Union (mIoU), mean class accuracy (mAcc), and per-class IoU, recall, and precision for grass, nest, and tree.
For a given scene and class \(c\), the metrics are defined as:

\[
\text{IoU}_c = \frac{TP_c}{TP_c + FP_c + FN_c}, \quad
\text{Rec}_c  = \frac{TP_c}{TP_c + FN_c},          \quad
\text{Prec}_c = \frac{TP_c}{TP_c + FP_c}.
\]

Scene-level mIoU and mAcc are computed by averaging IoU and recall, respectively, over the three classes for each scene.

For full transparency, we include a table with detailed per-scene metrics, covering \texttt{accuracy}, \texttt{miou}, \texttt{macc}, and per-class IoU, recall, and precision values for all three semantic categories.

\end{document}